\documentclass[pdflatex,sn-mathphys-num]{sn-jnl}

\usepackage{graphicx}%
\usepackage{multirow}%
\usepackage{amsmath,amssymb,amsfonts}%
\usepackage{amsthm}%
\usepackage{mathrsfs}%
\usepackage[title]{appendix}%
\usepackage{xcolor}%
\usepackage{textcomp}%
\usepackage{manyfoot}%
\usepackage{booktabs}%
\usepackage{algorithm}%
\usepackage{algorithmicx}%
\usepackage{algpseudocode}%
\usepackage{listings}%
\usepackage{comment}
\usepackage{float}
\usepackage{subcaption}
\usepackage{tabularx}
\DeclareMathOperator*{\argmax}{arg\,max}

\theoremstyle{thmstyleone}%
\theoremstyle{thmstyletwo}%

\theoremstyle{thmstylethree}%

\begin{document}

\title[Article Title]{Tracking Emotional Dynamics in Chat Conversations: A Hybrid Approach using DistilBERT and Emoji Sentiment Analysis}


\author[1]{\fnm{Ayan} \sur{Igali}}\email{a\_igali@kbtu.kz}

\author[1]{\fnm{Abdulkhak} \sur{Abdrakhman}}\email{a.abdulkhak@gmail.com}

\author[1]{\fnm{Yerdaut} \sur{Torekhan}}\email{e\_torekhan@kbtu.kz}

\author*[1]{\fnm{Pakizar} \sur{Shamoi}}\email{p.shamoi@kbtu.kz}

\affil*[1]{\orgdiv{School of Information Technology and Engineering}, \orgname{Kazakh-British Technical University}, \orgaddress{\street{Tole bi Street}, \city{Almaty}, \postcode{050000}, \country{Kazakhstan}}}


\abstract{

Computer-mediated communication has become more important than face-to-face communication in many contexts. Tracking emotional dynamics in chat conversations can enhance communication, improve services, and support well-being in various contexts. This paper explores a hybrid approach to tracking emotional dynamics in chat conversations by combining DistilBERT-based text emotion detection and emoji sentiment analysis. 
A Twitter dataset was analyzed using various machine learning algorithms, including SVM, Random Forest, and AdaBoost. We contrasted their performance with DistilBERT. 
Results reveal DistilBERT's superior performance in emotion recognition. Our approach accounts for emotive expressions conveyed through emojis to better understand participants' emotions during chats. 
We demonstrate how this approach can effectively capture and analyze emotional shifts in real-time conversations. Our findings show that integrating text and emoji analysis is an effective way of tracking chat emotion, with possible applications in customer service, work chats, and social media interactions.

}




\maketitle

\section{Introduction}\label{sec1}

Understanding and analyzing emotional cues is extremely important in the digital era, where communication increasingly occurs through text. 
The significance of digital communication, particularly chat-based interactions, is evident in various contexts. This capability has far-reaching benefits across various sectors, such as marketing, customer service, and mental health, leading to more empathetic and efficient engagements.


The shift to digital communication has transformed consumer behavior and interpersonal relationships. Several studies highlight the role of web-based chatting in this transformation \cite{ZINKHAN200317},  \cite{Peris2002}, with \cite{ZINKHAN200317}   focusing on the motivations and strategies of chatters, and \cite{Peris2002} emphasizing the complementarity of online and face-to-face relationships. However, \cite{Venter2019} raises concerns about the impact of computer-mediated communication on meaningful interpersonal interaction, particularly due to the lack of non-verbal cues and the potential for misunderstandings. In the organizational context, \cite{West2017} underscores the growing importance of social media in internal communication but also acknowledges the continued value of face-to-face interaction. Another work highlights the growing importance of social media as a customer service channel, emphasizing the need for its integration into customer interaction management tools \cite{Geierhos2011}.


Nowadays, chat systems enhance the user experience by employing graphics-based interfaces and including interactive elements such as avatars and emoticons \cite{bots}. Emojis play a significant role in digital communication, particularly in conveying humor and emotions \cite{Hasan2018}, \cite{Sampietro2020}. They help to bridge cultural differences and form digital communities, with their use varying based on the gender of the communicator \cite{Graham2019}. Emoticons, a type of emoji, are influenced by the social context and the relationship between communicators \cite{Derks2008}.


The ability to track emotional dynamics in chat conversations is crucial for understanding the evolution of interactions and relationships. Several studies emphasize the importance of this tracking \cite{Rybak2021}, \cite{Loglisci2018}, with \cite{Rybak2021} introducing a method for labeling emotional states over time and \cite{Loglisci2018} proposing a method for tracking collective emotional trajectories. \cite{Chen2018} and \cite{Ran2023} further contribute to this area by developing computational approaches for tracking and recognizing emotions in short text messages, with Chen's approach being particularly effective in real-time chat scenarios. These studies demonstrate the significance of tracking emotional dynamics in chat conversations for various applications, from qualitative discourse analysis to machine learning models.

The advent of Natural Language Processing (NLP) has transformed how 
machines comprehend human emotions, converting unprocessed text into usable data. Historically, traditional machine learning approaches such as Support Vector Machines (SVM) were used to analyze textual sentiments, employing feature extraction through Word2Vec techniques \cite{svmlib}. The recent shift towards more sophisticated models, especially with transformer-based architectures like DistilBERT, has significantly improved the understanding of context and subtleties in language\cite{distilbert}.

 This paper proposes a hybrid approach to evaluate emotional dynamics in chat conversations by fusing text and emoji emotion detection results. We focus on the importance of emojis in chat conversations, highlighting their expressive capability in digital communication. 
The contributions of this paper are as follows:
\begin{itemize}
    \item Comparative overview of ML and deep learning algorithms for emotion detection
    \item Introduction of the novel hybrid method for the fusion of emotions detected in text and emojis
\end{itemize}

The paper is structured as follows. Section I is this Introduction. Section 2 presents an overview of research works in chat emotion analysis. Section III describes methods, including data collection, text, and emoji emotion detection. Next, section IV presents experimental results. Finally, concluding remarks and future works are presented in Section V.

\section{Related Work}\label{sec2}

Sentiment analysis in NLP has progressed from basic lexicon-based methods to sophisticated techniques. Initially relying on word-based sentiment lexicons, the field transformed with machine learning, particularly through deep learning and transformer models like DistilBERT, which improved context understanding \cite{sailunaz2019}. This advancement is crucial for applications like social media analytics and customer feedback, highlighting its significance in modern NLP research.

 A range of studies have explored the dynamics of emotions in online chatting. The study \cite{chen} developed a computational approach to track and analyze emotions in real time, which was effective and efficient. Another study \cite{uvakov2014} used simulations to show that a Bot with a fixed emotion can influence the collective emotional state in a chat network. Next, \cite{Garas2012} found that emotional expressions in online chatrooms exhibit remarkable persistence, leading to specific emotional "tones" in these spaces. \cite{Mundra} proposed two weakly supervised approaches for detecting fine-grained emotions in contact center chat utterances, achieving high accuracy. Another study presented the emotion flow (EF) model in dialog systems, which incorporates memes \cite{memechat}.

Emotion recognition in conversations, also known as ERC,  is a branch of sentiment analysis, a growing field within NLP with important implications for various domains, including healthcare, education, human-computer interaction, and social sciences. Several ERC works employed emotion recognition from various sides. 

Several works investigated the application of emotion detection in chat conversations \cite{chat1, chat2, chat3}. One study presented the context encoder and emotion-shift detection model that can surpass chit-chat and even improve some aspects of task-oriented conversation \cite{chat1}. Another study commonly used techniques for ERC problems based on recent research on sequential, graph, and transformer models \cite{chat2}. Some works analyzed context modeling in conversations, speaker dependency, and methods for combining multimodal information \cite{chat4}.

One significant limitation of the reviewed methods is that they miss the emotional context provided by emojis. This can lead to an incomplete understanding of feelings and intentions in digital communication, as emojis are important in expressing emotions in chats. The analysis may miss important emotional nuances without emojis, resulting in less accurate data interpretation.

\section{Methods}

\subsection{Data}

\subsubsection{Dataset for Text Emotion Detection}
This study explores text-based emotion detection by analyzing two distinct datasets.

\paragraph{Dataset 1}

The first dataset \cite{saravia-etal-2018-carer}, derived from a collection of English Twitter messages, was used to identify six basic emotions: anger, fear, joy, love, sadness, and surprise.   It contains 20,000 data rows in 6 classes.

\begin{figure}[tb]
    \centering
    \begin{minipage}{0.48\textwidth}
        \centering
        \includegraphics[width=\linewidth]{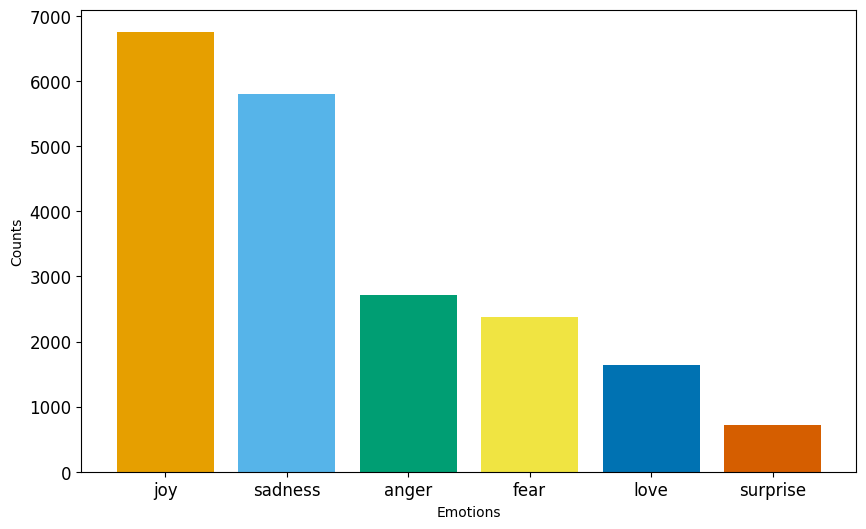}
        \caption{Aggregated emotion frequencies across the collected tweets.}
        \label{fig:twitter_dataset_bar_chart}
    \end{minipage}%
    \hspace{0.04\textwidth}%
    \begin{minipage}{0.48\textwidth}
        \centering
        \includegraphics[width=\linewidth]{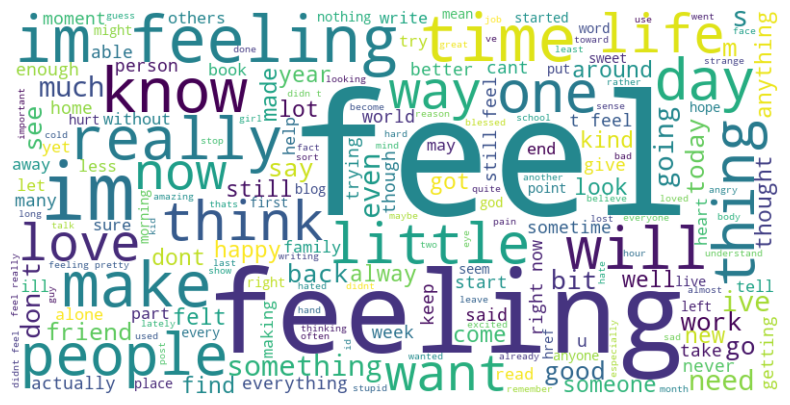}
        \caption{Word cloud generated from the Twitter dataset}
        \label{fig:wordcloudoftextdata}
    \end{minipage}
\end{figure}

As shown in Fig.~\ref{fig:twitter_dataset_bar_chart}, the aggregated emotion frequencies highlight significant trends in emotional expressions across the collected tweets. Furthermore, the word cloud in Fig.~\ref{fig:wordcloudoftextdata} provides a visual representation of the most frequently used words in the Twitter dataset, emphasizing the predominant themes and sentiments.

\paragraph{Dataset 2}


The second dataset was sourced from Kaggle and contains a collection of tweets annotated with emotions \cite{dataset2_kaggle}. It contains 13 classes and has a size of 39,774 data rows. This dataset presents a significant challenge due to the imbalanced nature of emotion representation and the number of emotional states, ranging over thirteen different sentiments.

\begin{figure}[tb]
    \centering
    \begin{minipage}{0.48\textwidth}
        \centering
        \includegraphics[width=\linewidth]{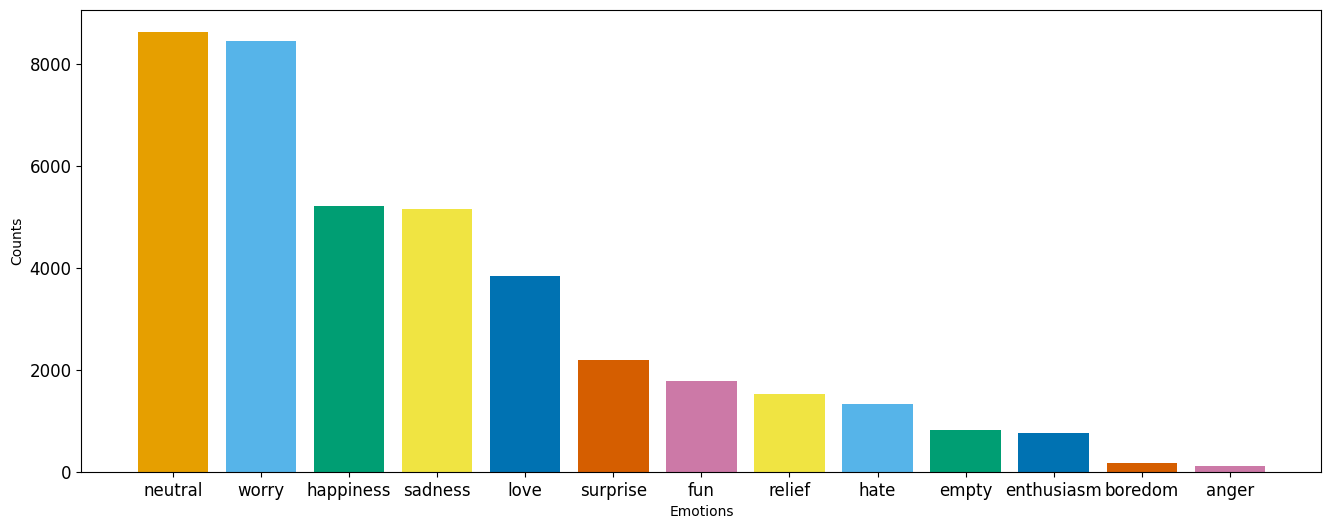}
        \caption{Bar chart illustrating the frequency of each emotion in the dataset}
        \label{fig:secondtwitterbarchart}
    \end{minipage}\hspace{0.04\textwidth}%
    \begin{minipage}{0.48\textwidth}
        \centering
        \includegraphics[width=\linewidth]{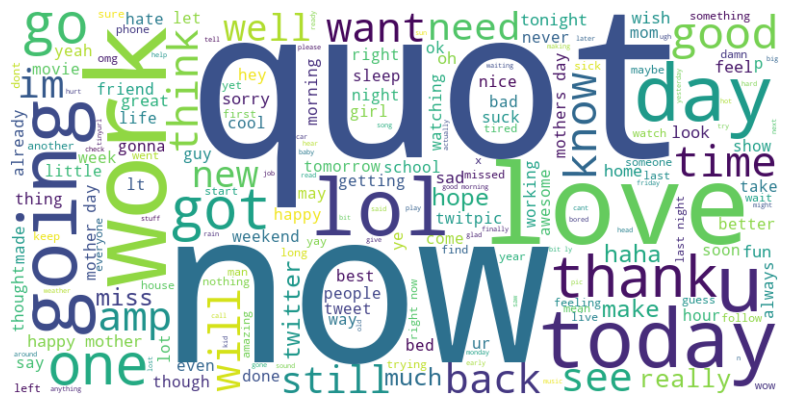}
        \caption{Word cloud representing the linguistic patterns associated with various emotions in the Kaggle dataset}
        \label{fig:secondtwitterwordcloud}
    \end{minipage}
\end{figure}

\subsubsection{Dataset for Emoji Emotion Detection}

We used the Emoji Sentiment Ranking dataset \cite{esr}. It provides standardized sentiment scores for various emojis commonly employed in digital communication. As summarized in Table \ref{tab:emoji_stats}, the dataset encompasses 751 instances each for negative, neutral, and positive sentiments. The mean sentiment scores illustrate a predominant positivity with a value of $\approx 0.447$, despite the range of scores spanning from a minimum of $\approx$ 0.007 to a maximum of $\approx 0.972$ for positive sentiments. For example, for the emoji "SMILING FACE WITH HEART-SHAPED EYES" (Unicode codepoint 0x1f60d), the negative, neutral, positive, and compound sentiment scores are 0.052, 0.219, 0.729, and 0.678, respectively.

\begin{table}[tb]
\centering
\caption{Statistical Summary of Emoji Sentiment Scores}
\label{tab:emoji_stats}
\begin{tabular}{|c|c|c|c|c|c|c|c|c|}
\hline
\textbf{Sentiment} & \textbf{Count} & \textbf{Unique}  & \textbf{Freq} & \textbf{Mean} & \textbf{Max} & \textbf{Min} \\ \hline
Neg & 751 & 271  & 31 & 0.164 & 0.778 & 0.006 \\ \hline
Neut & 751 & 360  & 30 & 0.389 & 0.987 & 0.014 \\ \hline
Pos & 751 & 374  & 24 & 0.447 & 0.972 & 0.007 \\ \hline
\end{tabular}
\end{table}

\begin{figure}[H]
    \centering
    \includegraphics[width=\linewidth]{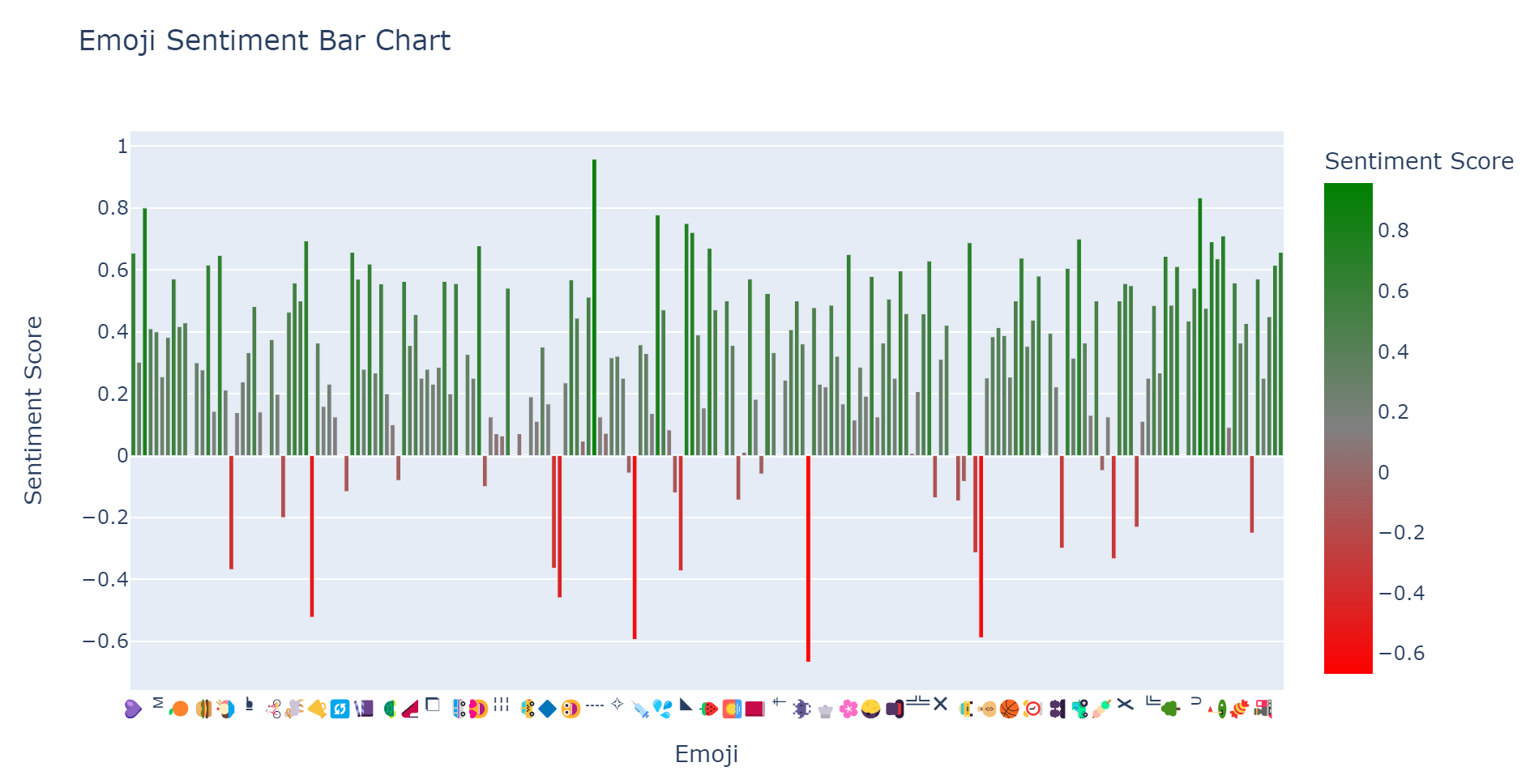}
    \caption{A bar chart representing the sentiment scores of various emojis, based on the standardization by \cite{esr}}
    \label{fig:emojigraph}
\end{figure}

Fig.~\ref{fig:emojigraph} displays a bar chart of the sentiment scores for various emojis \cite{esr}. This visualization provides a clear and quantifiable overview of how different emojis are typically interpreted in terms of sentiment.

This dataset is critical for validating the effectiveness of emotion recognition and understanding emojis' role in chat communications. Evaluation of how emojis contribute to emotional expression in text is crucial for enhancing automatic emotion detection systems in practical scenarios \cite{chat1, distilbert, chat2}.

\subsubsection{Synthetic Work Chat Dataset}

\begin{figure}[tb]
    \centering
    \includegraphics[width=0.8\linewidth]{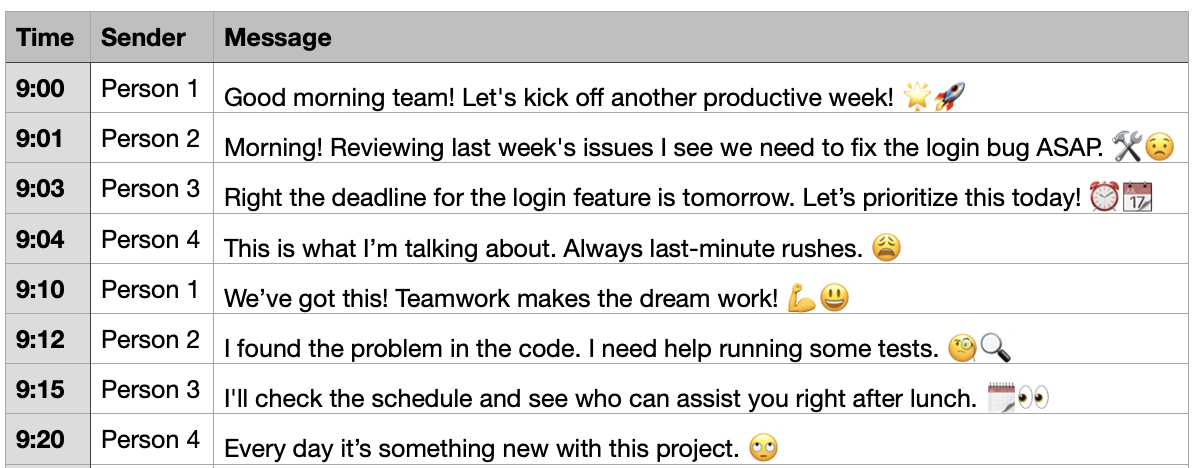}
    \caption{Examples of records in the collected dataset showcasing the interaction among team members over a year-long project.}
    \label{fig:chatimage}
\end{figure}

Messages in this dataset were artificially generated by ChatGPT 4.0 to simulate a realistic project-focused group chat environment involving four participants: Person 1, Person 2, Person 3, and Person 4. As illustrated in Fig. \ref{fig:chatimage}, the dataset encompasses 113 chat messages for 1 full work day from 9 a.m. to 6 p.m. We will use it to analyze emotional dynamics in a professional communication setting.

Messages reflect how team members exchange insights and updates about the project's progress and potential issues. In the simulated dialogues, each participant adopts a distinct role and emotional tone, employing emojis to enhance the expressiveness of their messages.

Key interactions in the dataset include:
\begin{itemize}
\item Person 1: Motivator and encourager, initiates optimistic discussions and frequently encourages the team.
\item Person 2: Technical expert with a cynical edge.  Raises technical and procedural concerns.
\item Person 3: Taskmaster and coordinator who is focused on deadlines and deliverables, managing the project timeline, and coordinating tasks.

\item Person 4: Realist and expresses urgency. Expresses frustration and urgency about the project, highlighting issues and challenges.

\end{itemize}

\begin{figure}[tb]
    \centering
    \includegraphics[width=0.9\linewidth]{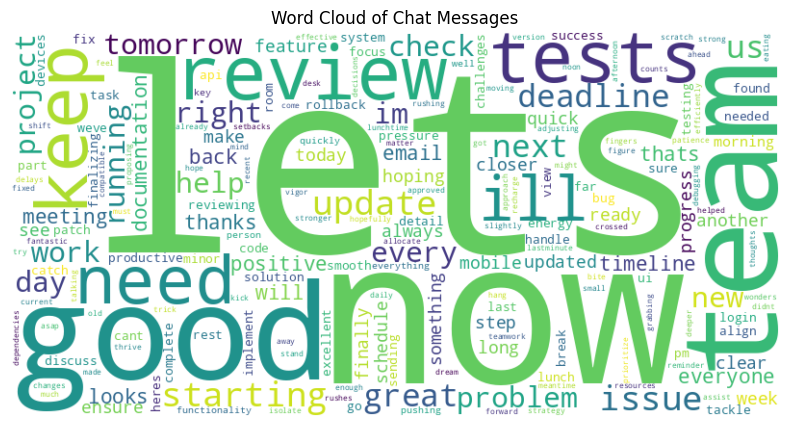}
    \caption{Word cloud depicting the most frequently used words in the synthetic dataset}
    \label{fig:wordcloud}
\end{figure}

This artificial dataset supports the analysis of how emotions are conveyed through text and emojis in professional exchanges and serves as an example for validating the effectiveness of emotion recognition algorithms under varied communicative contexts \cite{chat1, chat2}. Fig. \ref{fig:wordcloud} visualizes the dataset's most frequently used words.
 
ional settings \cite{bhavani2021, distilbert}.

\subsection{Text Emotion Detection}

Analysis of emotions can be derived from various channels, such as colors \cite{color}, music \cite{music}, faces \cite{face}, \cite{text_emotion}, multichannel \cite{10585521,movie}, emoji \cite{Srivastava2021}. Each channel provides unique data, allowing for a more detailed understanding of emotional states.

\subsubsection{Machine Learning Algorithms}
\paragraph{SVM}

Support Vector Machines (SVM) is a supervised, machine learning algorithm commonly used for classification tasks \cite{svmlib}. The algorithm behind SVMs classifies input data $\boldsymbol{x}$ by finding the optimal hyperplane that maximizes the distance between opposite classes $\boldsymbol{y}$ in the feature space. 

For training sample $\boldsymbol{x_i} \in R^n, i=1,..., m, $ which given in two classes, and label vector $\boldsymbol{y} \in R^m$ such that $y_i \in \{-1, 1\}$, SVM solves the following optimization problem:

\begin{equation}
    \frac{1}{2} \omega^T\omega + C \sum_{i=1}^m \xi_i \rightarrow \min_{\omega, b, \xi}
\end{equation}

\begin{equation}
    \text{subject to} \; y_i(\omega^T \phi(x_i) +b) \ge 1 -\xi_i, \;
    \xi_i \ge 0, i = 1, ..., m,
    \label{solved_problem}
\end{equation}
where $\phi(x_i)$ maps into higher-dimensional space and $C > 0$ is the regularization parameter.

\paragraph{GaussianNB}

Naive Bayes is a probabilistic and supervised machine learning algorithm used for classification tasks \cite{naiveb}. It is based on Bayes` theorem and makes a  ``naive`` assumption about the independence of features. For classification, it uses:

\begin{equation}
    P(y|X) = \frac{P=(X|y)P(y)}{P(X)}
\end{equation}

The Naive Bayes classifier picks the most probable hypothesis, which is known as the maximum a posteriori.

\begin{equation}
    \hat{y} =  {argmax}_y P(y) \prod^n_{i=1}P(x_i|y)
\end{equation}

Gaussian Naive Bayes assumes that the likelihood of feature values associated with each class follows a normal (or Gaussian) distribution.

\paragraph{AdaBoost}

Adaptive Boosting (AdaBoost) is a supervised machine learning algorithm where a classifier is built incrementally \cite{adaboost}. Each iteration employs a basic learning algorithm called the base learner, producing a classifier. These classifiers are then assigned weight coefficients. Ultimately, the final classification is determined through a weighted combination of the base classifiers' decisions.
\paragraph{Decision Tree}

\begin{figure}[!h]
    \centering

    \includegraphics[width=0.9\linewidth]{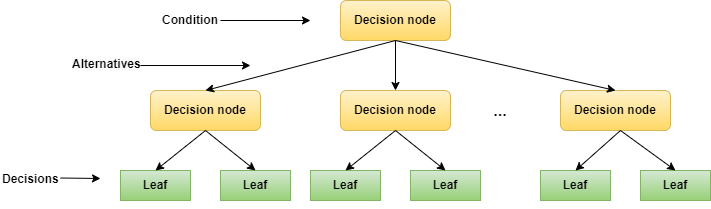}
    \caption{Schematic representation of decision tree algorithm. }
    \label{fig:decisiontree_schema}
\end{figure}

Decision Tree is a non-parametric, supervised machine learning algorithm for classification and regression tasks \cite{decisiontree}. It adopts a hierarchical tree structure with root, internal and leaf nodes, and branches (see Fig. \ref{fig:decisiontree_schema}). There are two methods of choosing the best attributes, information gain and Gini impurity, that act as splitting criteria for decision trees.

\paragraph{K-Nearest Neighbors}

K-Nearest Neighbors (KNN) is a simple, non-parametric, supervised machine learning algorithm for classification and regression tasks \cite{mlmurphy}. In classification tasks, the algorithm calculates the distance between input data $\boldsymbol{x}$ and training samples $\boldsymbol{X}$, then determines class $\boldsymbol{\hat{y}}$ of input data by the nearest $\boldsymbol{k}$ neighbors in the feature space. Distance between data points can be calculated by Euclidean (\ref{euclidean}), Manhattan (\ref{manhattan}) or Minkowski (\ref{minkowski}) distances:
\begin{equation}
d(x, y) = \sqrt{\sum_i{(y_i - x_i)^2}}
\label{euclidean}
\end{equation}
\begin{equation}
d(x, y) = \sum_i \vert x_i - y_i \vert
\label{manhattan}
\end{equation}
\begin{equation}
d(x, y) = \left(\sum_i \vert x_i - y_i\vert^p\right)^{1/p}
\label{minkowski}
\end{equation}

The class $\boldsymbol{\hat{y}}$ of input data $\boldsymbol{x}$ is set to the most common class among $\boldsymbol{k}$ neighbours of $\boldsymbol{x}$:
\begin{equation}
    \hat{y} = argmax \sum^k_i{I[y = y_{(i)}]}
\end{equation}
\paragraph{Random Forrest}

\begin{figure}[!h]
    \centering
    \includegraphics[width=0.9\linewidth]{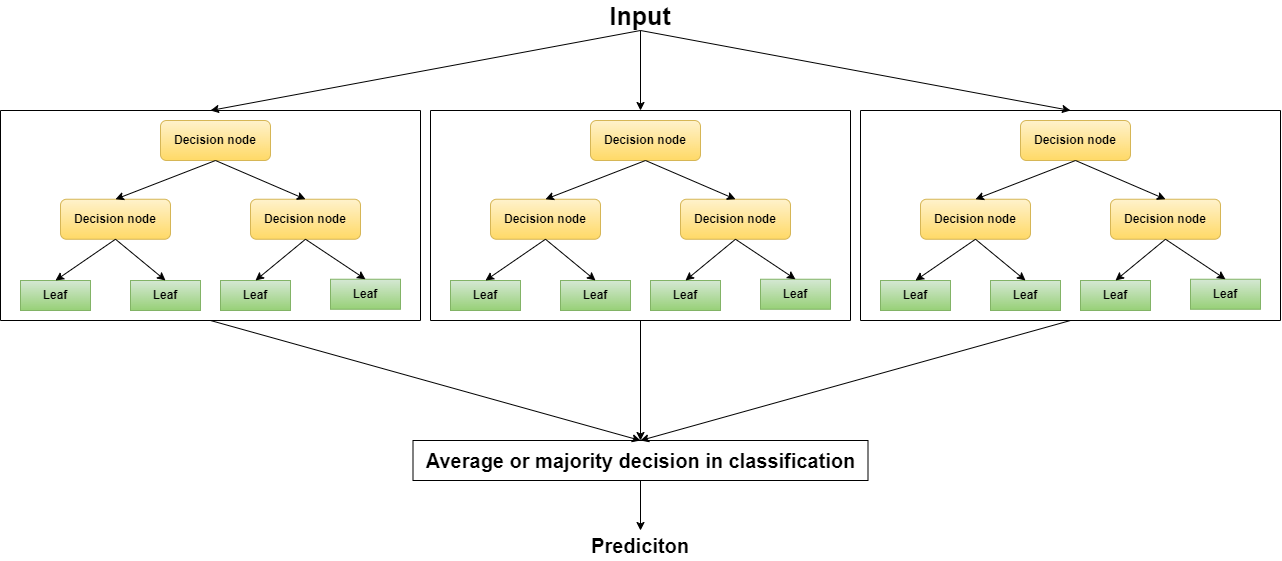}
    \caption{Schematic representation of Random Forest algorithm.}
    \label{fig:randomforest_schema}
\end{figure}

Random forests are supervised ensemble learning algorithms used for tasks such as classification and regression \cite{mlmurphy}. During training, they generate multiple decision trees (see Fig. \ref{fig:randomforest_schema}). For classification tasks, the overall prediction is based on the most frequently selected class among these trees.

\subsubsection{Deep Learning Approach}
\begin{figure}
    \centering
    \includegraphics[width=0.9\linewidth]{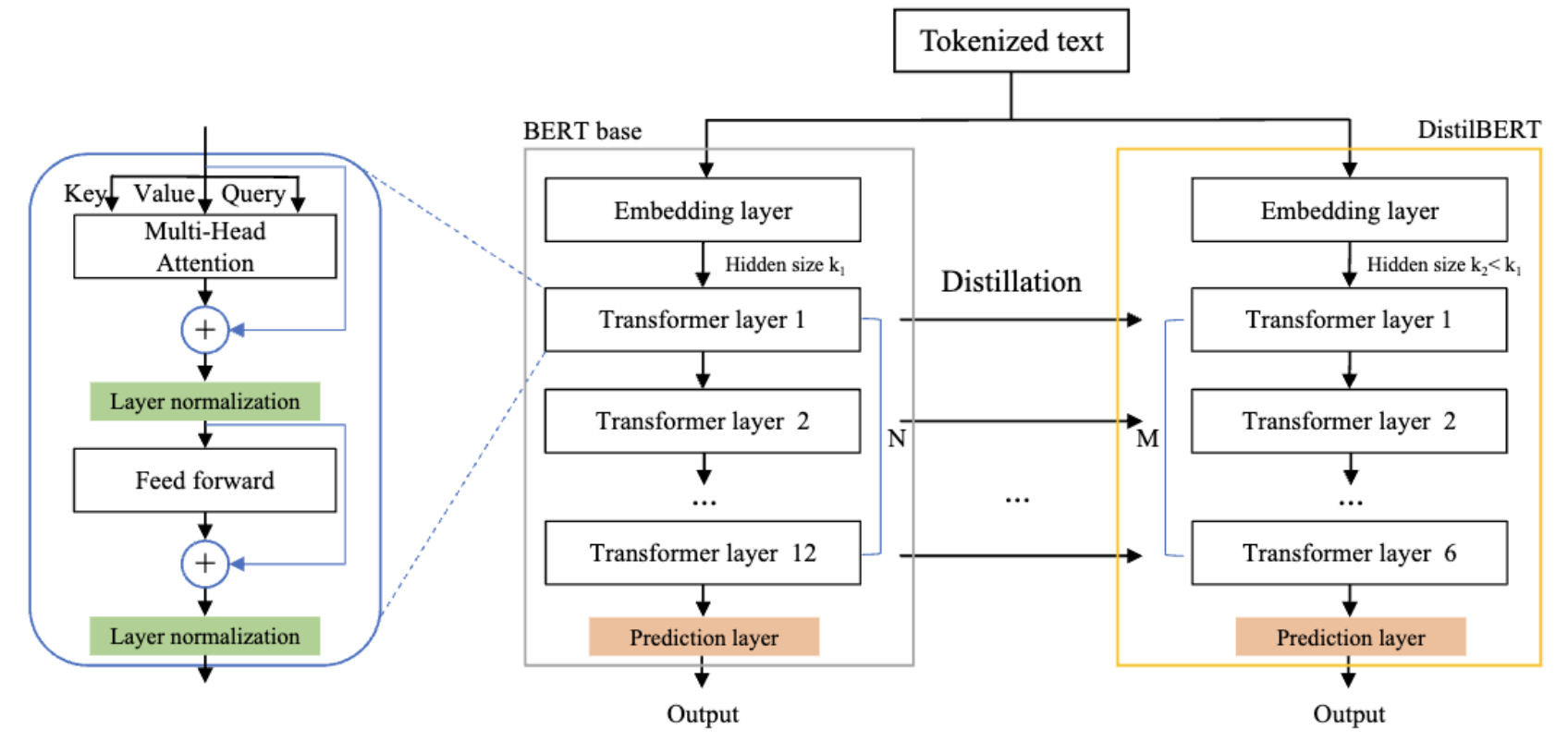}
    \caption{The DistilBERT model architecture \cite{dbert}}
    \label{fig:enter-label}
\end{figure}

DistilBERT \cite{distilbert}, a streamlined version of BERT (Bidirectional Encoder Representations from Transformers), is designed to be smaller, faster, and more efficient while retaining most of BERT's performance. It uses the same foundational transformer architecture and training objectives as BERT, with some modifications for efficiency (see Fig. \ref{fig:enter-label}).
The DistilBert self-attention mechanism computes attention scores based on the input vectors (Eq. \ref{eq:attention}):

\begin{equation}
    \text{Attention}(Q, K, V) = \text{softmax}\left(\frac{QK^T}{\sqrt{d_k}}\right) V
    \label{eq:attention}
\end{equation}

Where $Q$, $K$, and $V$ represent the query, key, and value matrices, respectively, and $d_k$ is the dimensionality of the key vectors.
Next, Multi-head attention allows the model to attend to information from different representation subspaces at different positions (\ref{eq:multihead}, \ref{eq:head}):
\begin{equation}
    \text{MultiHead}(Q, K, V) = \text{Concat}(\text{head}_1, \dots, \text{head}_h)W^O
    \label{eq:multihead}
\end{equation}

\begin{equation}
    \text{head}_i = \text{Attention}(QW_i^Q, KW_i^K, VW_i^V)
    \label{eq:head}
\end{equation}

Here, $W_i^Q$, $W_i^K$, $W_i^V$, and $W^O$ are parameter matrices.
Each transformer block in DistilBert contains a feed-forward network, which applies the following transformations (\ref{eq:ffn}):
\begin{equation}
    \text{FFN}(x) = \text{max}(0, xW_1 + b_1)W_2 + b_2
    \label{eq:ffn}
\end{equation}

To help stabilize the learning process, techniques such as layer normalization and residual connections are used (\ref{eq:layernorm}):

\begin{equation}
    \text{LayerNorm}(x + \text{Sublayer}(x))
    \label{eq:layernorm}
\end{equation}

where $\text{Sublayer}(x)$ could be either an attention mechanism or a feed-forward network.
DistilBERT uses a distillation loss, often parameterized by a temperature-scaled cross-entropy loss, to help the smaller model learn from the larger model.

\subsection{Emoji Emotion Detection}
In digital communication, emojis are acknowledged to significantly enhance the conveyance of emotional nuance.

\begin{figure}[tb]
    \centering
    \includegraphics[width=\linewidth]{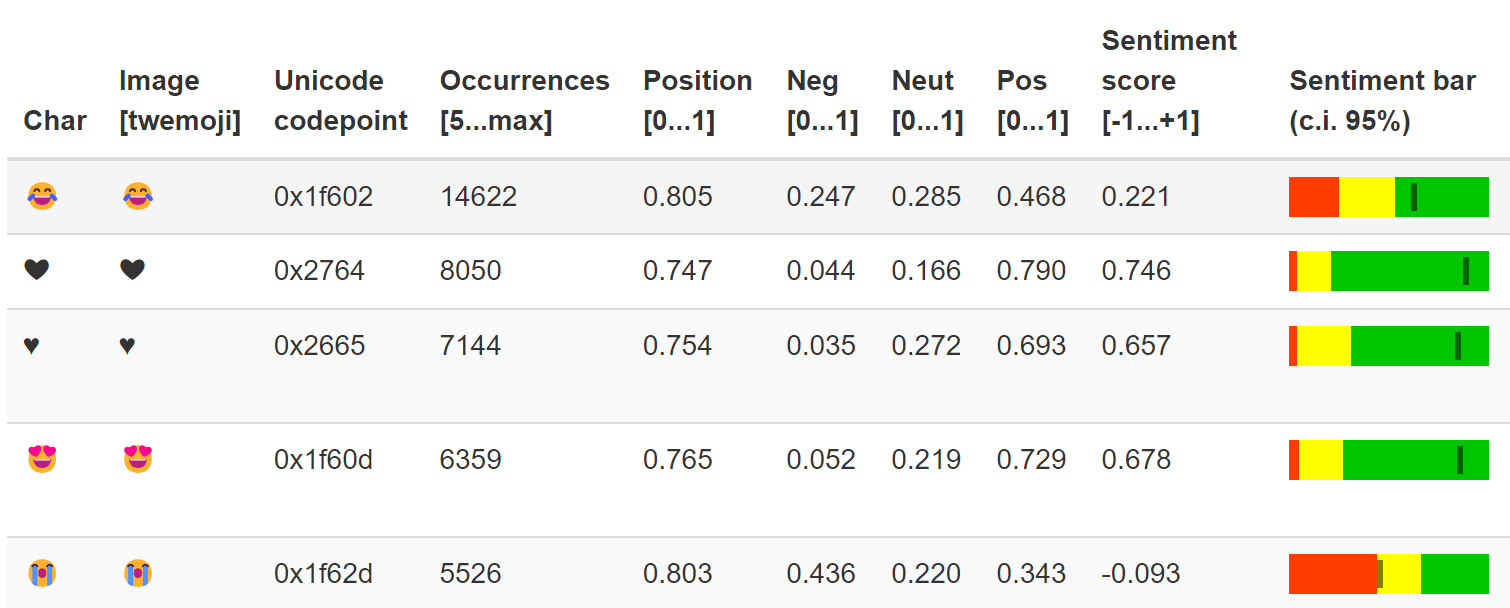}
    \caption{Emoji Sentiment Ranking from \cite{esr}, showing the distribution of sentiment scores among popular emojis}
    \label{fig:emoji_sentiment_ranking}
\end{figure}

A sentiment lexicon, which attributes sentiment scores to a broad spectrum of emojis, has been employed for emoji-based emotion detection. This ranking categorizes emojis by the intensity and polarity of the sentiments they typically express, with scores derived from user feedback \cite{articleKralj}. As demonstrated in Fig. \ref{fig:emoji_sentiment_ranking}, the Emoji Sentiment Ranking offers a comprehensive visual overview of sentiment scores for commonly used emojis.

Emojis are extracted from a corpus of messages, and each emoji is correlated with its associated emotional sentiment according to the lexicon. The collective presence of emojis within messages is interpreted as a composite expression, contributing to the overall sentiment. This method integrates emoji sentiment analysis into the emotion detection framework, substantiating the role of emojis as a unique communicative lexicon.

\subsection{Proposed Approach}

Our approach focuses on analyzing chat conversations by integrating NLP with emoji sentiment scoring, as seen in Fig. \ref{flow}. This section presents the mathematical framework that supports feature extraction, sentiment analysis, and intensity adjustment:

\begin{figure}[tb]
\centering
\includegraphics[width=\textwidth]{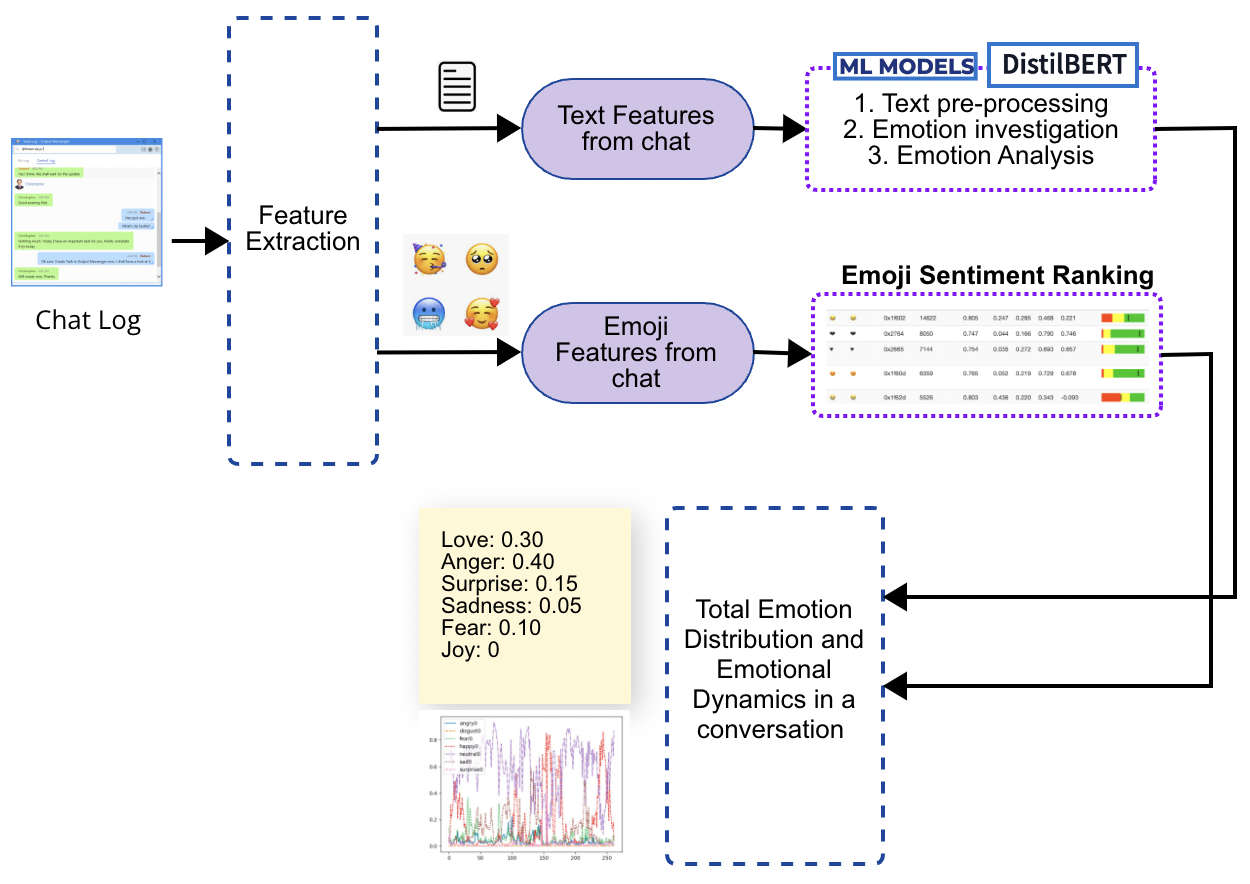}
\caption{Schematic representation of the proposed hybrid approach}
\label{flow}
\end{figure}

\begin{itemize}
    \item \textbf{Feature Extraction.} Textual features are derived using a tokenizer function $\tau$ from the pre-trained DistilBERT model, transforming text into a sequence of tokens $\mathbf{T}$. Meanwhile, emoji features are identified through a set of regular expressions $\mathcal{R}$, which capture the Unicode representations of emojis $\mathcal{E}$ from the text.
    \item \textbf{Intensity Factor for Emoji Sentiment Analysis.} For each emoji $e$, the sentiment intensity factor $\phi(e)$ is determined from the emoji sentiment ranking $\mathcal{S}$. The ranking assigns positive $p_i$, negative $n_i$, and neutral $u_i$ sentiment scores to each emoji $e_i \in \mathcal{E}$, where $\mathcal{S} = \{s_1, s_2, \ldots, s_k\}$, and $s_i = (p_i, n_i, u_i)$. The intensity factor $\phi(e_i)$ for each emoji is computed as follows:
\begin{equation}
    \phi(e_i) = 1 + \max(p_i, n_i, u_i)
\end{equation}
\item \textbf{Sentiment Analysis with DistilBERT}. The base emotional score is obtained using the DistilBERT model $\mathcal{D}$, which produces a set of logits $\mathcal{L}(x)$ for given text $x$. These logits are processed through a softmax function $\sigma$ to yield a probability distribution over possible emotions $E$:
\begin{equation}
    P(E|x) = \sigma(\mathcal{L}(x)) = \frac{e^{\mathcal{L}(x)}}{\sum_{j} e^{\mathcal{L}(x_j)}}
\end{equation}
The emotion prediction $\hat{e}$ and its corresponding textual sentiment score $\psi(x)$ are determined by:
\begin{equation}
    \hat{e} = \argmax_{e \in E} P(E|x)
\end{equation}

\begin{equation}
    \psi(x) = \max_{e \in E} P(E|x)
\end{equation}
\item \textbf{Integration of Text and Emoji Sentiments}. The final sentiment score $\Omega(x)$ for text $x$ combines the textual sentiment score $\psi(x)$ and the emoji sentiment intensity factors $\Phi(x)$. The computation of $\Omega(x)$ is as follows:
\begin{equation}
    \Omega(x) = \psi(x) \times \left(\frac{1}{|\Phi(x)|}\sum_{\phi \in \Phi(x)} \phi \right)
\end{equation}

Where each $\phi \in \Phi(x)$ represents the sentiment intensity contributed by an individual emoji. The expression $\frac{1}{|\Phi(x)|}\sum_{\phi \in \Phi(x)} \phi$ calculates the average sentiment intensity of all the emojis in the text.  It scales the textual sentiment score $\psi(x)$ based on the average sentiment intensity of the emojis. This factor is always greater than or equal to 1, ensuring that the final sentiment score $\Omega(x)$ is at least as strong as the textual sentiment score. The final score $\Omega(x)$ reflects the overall sentiment by integrating both the text and the emojis.

Note that in the case  of just one emoji, the equation is as follows:
\begin{equation}
    \Omega(x) = \psi(x) \times  \phi
    \label{emoji1}
\end{equation}

\end{itemize}

\section{Results}
\begin{table*}
\centering
\resizebox{\linewidth}{!}{%
\begin{tabular}{l|cccc|cccc}
\toprule
\textbf{Method} & \multicolumn{4}{c}{\textbf{Dataset 1 (6 classes, size - 20,000)}} & \multicolumn{4}{c}{\textbf{Dataset 2 (13 classes, size - 39,774 )}} \\
\midrule
 & \textbf{Accuracy} & \textbf{Precision} & \textbf{Recall} &\textbf{F1} & \textbf{Accuracy}  & \textbf{Precision} & \textbf{Recall} &  \textbf{F1} \\
\hline
SVM           & 0.20 & 0.21 & 0.20 & 0.16 & 0.17 & 0.18 & 0.17 & 0.16 \\
GaussianNB    & 0.19 & 0.19 & 0.19 & 0.16 & 0.12 & 0.13 & 0.12 & 0.10 \\
AdaBoost      & 0.26 & 0.25 & 0.26 & 0.24 & 0.16 & 0.15 & 0.16 & 0.15 \\
Decision Tree & 0.75 & 0.72 & 0.75 & 0.73 & 0.79 & 0.76 & 0.79 & 0.77 \\
KNN           & 0.67 & 0.67 & 0.67 & 0.67 & 0.75 & 0.71 & 0.75 & 0.72 \\
RandomForest  & 0.78 & 0.76 & 0.78 & 0.77 & 0.81 & 0.79 & 0.82 & 0.80 \\
DistilBERT    & 0.93 & 0.90 & 0.91 & 0.90 & 0.38 & 0.27 & 0.20 &  0.19\\
\bottomrule
\end{tabular}%
}
\caption{Comparison of various machine learning algorithms and DistilBERT across two datasets using Accuracy, Precision, Recall, and F1 Score as performance metrics.}
\label{tab:comparison_metrics}
\end{table*}

\subsection{Performance Evaluation}

\begin{figure}[tb]
    \centering
    \includegraphics[width=1\textwidth]{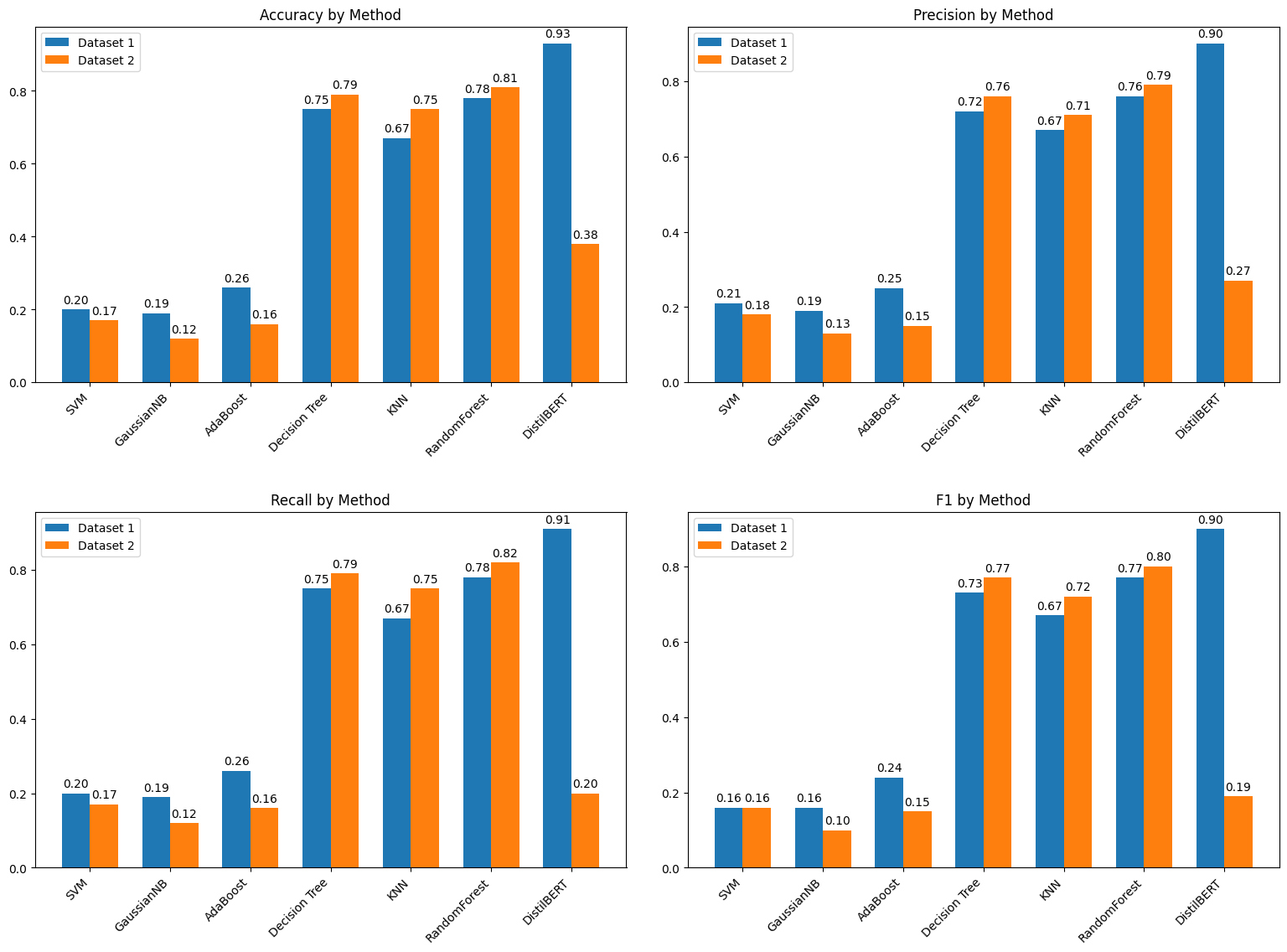}
    \caption{Illustration of metrics of ML models and Distilbert performance as bar charts}
    \label{ml_metics}
\end{figure}

Now let us compare the performance of various machine learning models and DistilBERT across two datasets using Accuracy, Precision, Recall, and F1 Score as performance metrics. Our findings show that performance varied significantly across different models (see Table \ref{tab:comparison_metrics}, Fig. \ref{ml_metics}). 


For Dataset 1 (6 classes, 20,000 instances), SVM and GaussianNB show low performance, with metrics around 0.20 and 0.19, respectively. AdaBoost performs slightly better, with an Accuracy of 0.26. Decision Tree and KNN show good performance, with Accuracy of 0.75 and 0.67, respectively. RandomForest exhibits excellent performance, with an Accuracy of 0.78. DistilBERT outperforms all models with an Accuracy of 0.93, Precision of 0.90, Recall of 0.91, and F1 Score of 0.90.

For Dataset 2 (13 classes, 39,774 instances), SVM and GaussianNB again show poor performance, with GaussianNB having the lowest metrics around 0.12. AdaBoost also performs poorly, with an Accuracy of 0.16. Decision Tree and KNN perform well with an Accuracy of 0.79 and 0.75, respectively. RandomForest achieves the highest performance among traditional models with an Accuracy of 0.81. DistilBERT shows relatively low performance with an Accuracy of 0.38, Precision of 0.27, Recall of 0.20, and F1 Score of 0.19. Notably, ensemble methods such as Random Forest demonstrated superior performance. This variance in performance highlights the complexity and diversity of emotion detection tasks within digital communication environments.

In summary, DistilBERT excels in Dataset 1 but is outperformed by Decision Tree and RandomForest in Dataset 2. RandomForest and Decision Tree consistently perform well across both datasets, while SVM, GaussianNB, and AdaBoost exhibit lower performance.

\subsection{Analysis of Emotional Dynamics within Professional Communication}
\label{subsec:emotional_dynamics}

This subsection explores emotional dynamics within a professional communication setting using the proposed approach. The goal is to analyze the distribution and track the changes in emotional expressions among team members during one working day's interactions in a chat.

\begin{figure}[tb]
    \begin{subfigure}{0.49\textwidth}
        \centering
        \includegraphics[width=1\linewidth]{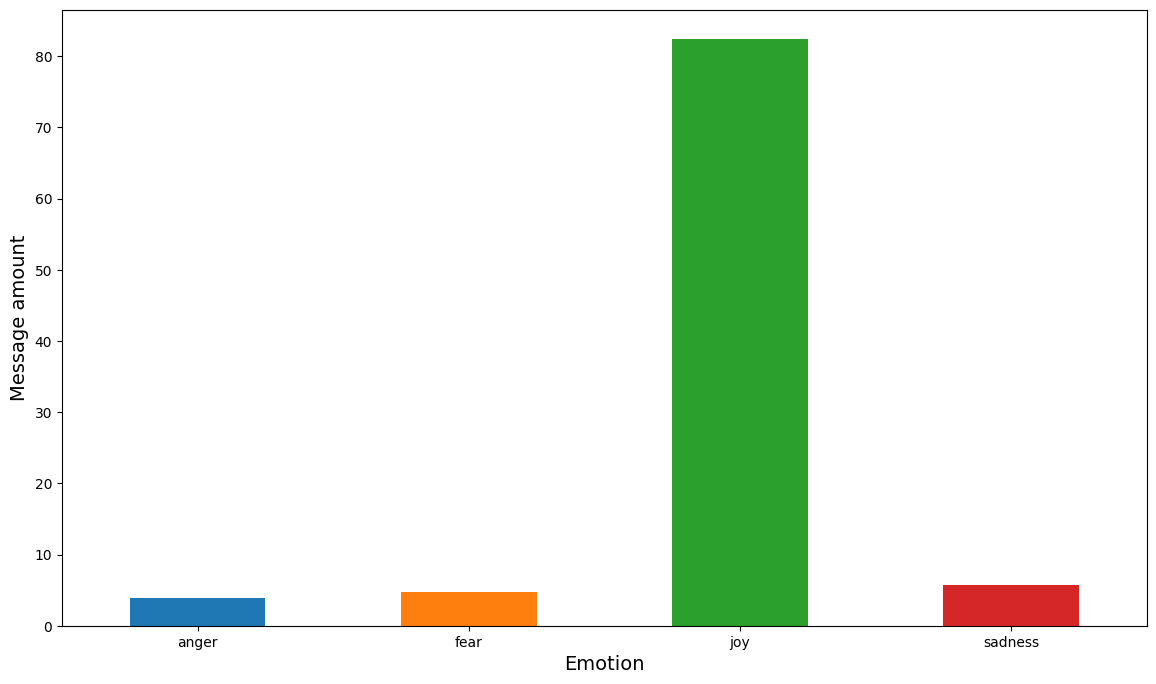}
           \caption{Total amount of emotions}
              \label{chat_barcharts1}
    \end{subfigure}
    \begin{subfigure}{0.49\textwidth}
        \centering
        \includegraphics[width=1\linewidth]{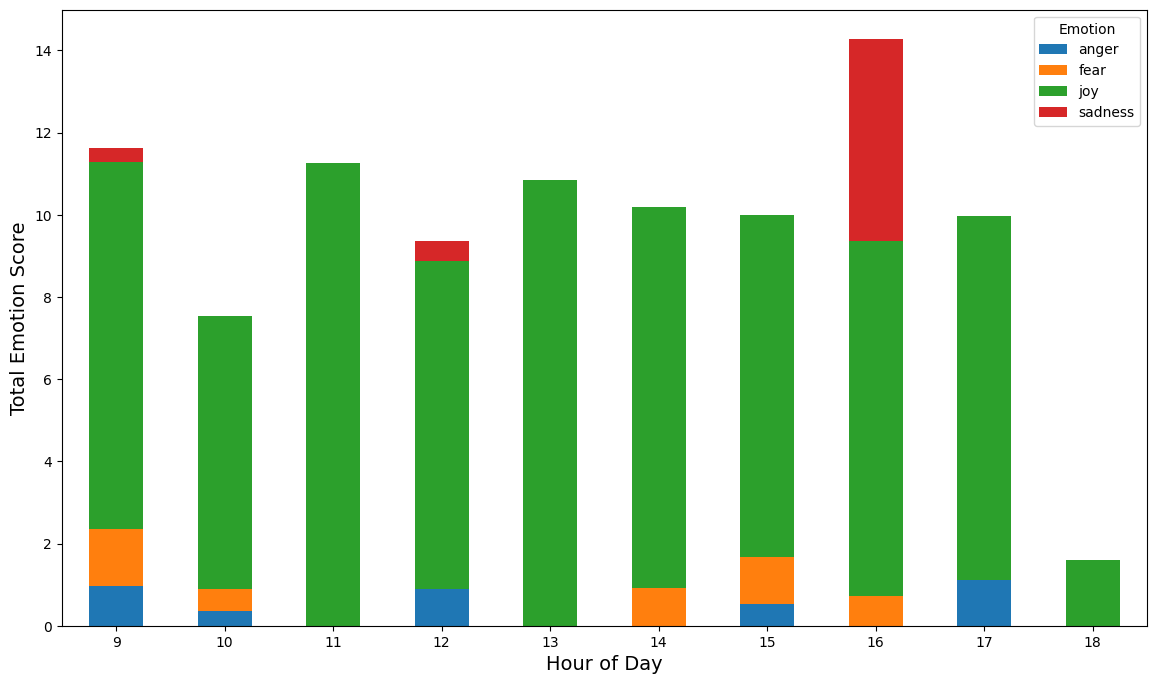}
           \caption{Distribution by hours during one day}
             \label{chat_barcharts2}
    \end{subfigure}
    \caption{Participants' most frequently exhibited emotions}
    \label{chat_barcharts}
\end{figure}

Fig. \ref{chat_barcharts1} and Fig. \ref{chat_barcharts2} illustrate the distribution of dominant emotions exhibited in the chat messages. As shown, \textit{joy} is the most prevalent emotion, followed by \textit{sadness}. \textit{Anger} and \textit{fear} are also present but in fewer messages. No messages with the dominant emotions of \textit{love} and \textit{surprise} were found. Fig. \ref{chat_barcharts2} illustrates stacked emotion scores over a full working day. It can be seen that the emotion of \textit{sadness} was abnormally high towards the end of the day, possibly because some goals planned for the day were not achieved. However, the emotion of \textit{joy} was dominant throughout the day, likely because people used polite and greeting words and tried to maintain a positive atmosphere.

\begin{figure}[H]
    \begin{subfigure}{0.49\textwidth}
        \centering
        \includegraphics[width=0.9\textwidth]{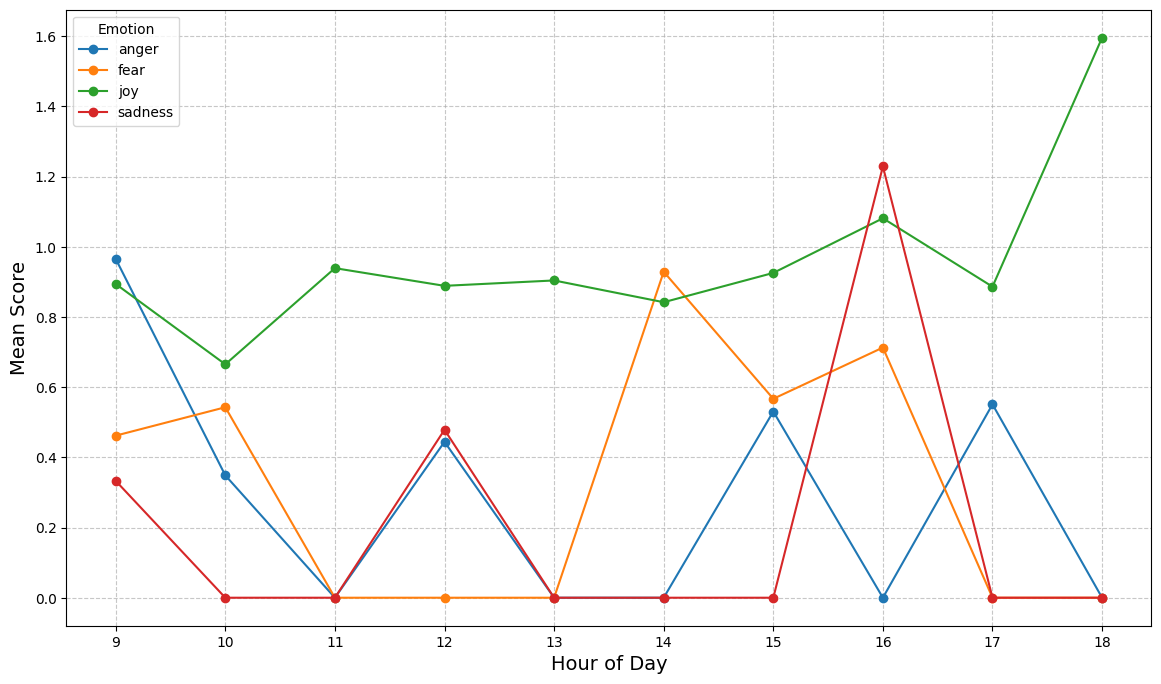}
      \caption{Emotion tracking results using Mean score}
             \label{fig:line_chart1}
    \end{subfigure}
    \begin{subfigure}{0.49\textwidth}
        \centering
        \includegraphics[width=0.9\textwidth]{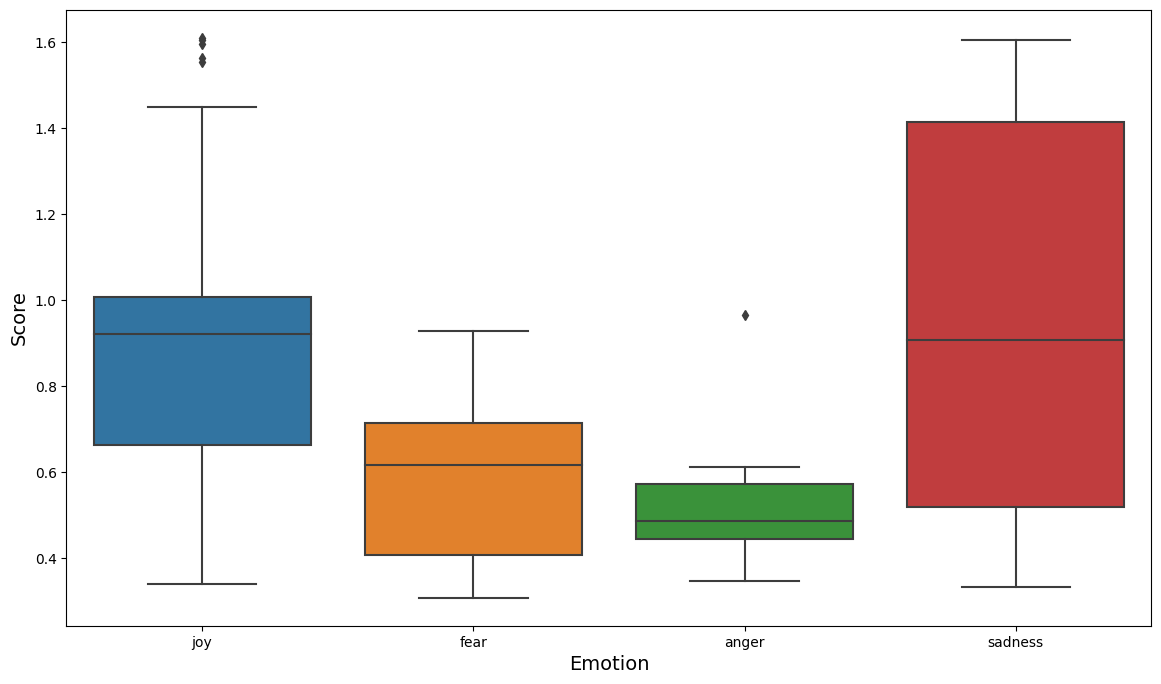}
    \caption{Distribution of scores for each emotion}
        \label{fig:line_chart2}
    \end{subfigure}
   \caption{Emotion tracking results}
    \label{fig:line_chart}
\end{figure}

 Next, Fig. \ref{fig:line_chart1} illustrates the emotion tracking results in the analyzed chat. It can be seen that \textit{joy} was dominant throughout the day, but \textit{fear} and \textit{sadness} became noticeable towards the end of the day, possibly due to the pressure of meeting deadlines. Fig. \ref{fig:line_chart2} shows the distribution of scores for each emotion and their respective intensities. It is evident that the intensity of \textit{sadness} is significantly higher than other emotions.

In conclusion, the analysis revealed the complex emotional interactions in professional settings, highlighting the various emotions people experience in collaborative environments. These findings are important for creating supportive workplaces, understanding how people interact, and improving communication to boost team productivity and emotional well-being \cite{chat1, chat2, chat3}.

The proposed emotion tracking approach has various applications, such as:
\begin{itemize}
    \item Social Media Analysis. Understanding public emotional trends on social media platforms posts involving conversations.
    \item Education. Improving online learning environments by responding to students' emotional states. This can involve face emotion recognition as an additional channel.
    \item Customer Support Services. Improving company response strategies by analyzing customer emotions during interactions.
    \item Chatbots. Developing more empathetic AI chatbots and virtual assistants.
    \item Tracking mental health. For example, in support groups or therapy through chat.
\end{itemize}

\section{Conclusion}


This paper uses advanced text analysis and emoji sentiment mapping to focus on emotion and sentiment detection in chat conversations. It compares traditional ML methods and the DistilBERT transformer model in the text's context of emotion detection. The results highlight the superior accuracy and F1 scores achieved by DistilBERT, demonstrating the significant impact of deep learning in NLP. Moreover, incorporating emoji analysis has been crucial in bridging the gap between textual data and emojis' rich, emotive context.

The study has several limitations. Firstly, the analysis relied on synthetically generated chat data rather than real-world conversations, which may limit the generalizability of our findings. Secondly, while we used emoji sentiment analysis, analyzing the specific emotions conveyed by emojis would be more accurate, as they can represent a wide range of emotions that sentiment analysis may not capture.

Future work should address these limitations by incorporating real-world chat data. Building a fully functional application that can track and analyze emotional dynamics in real-time chat conversations will be a significant next step, allowing for practical applications in various domains. Additionally, expanding the analysis to include multimodal data analysis, such as images or voice messages, could provide a better understanding of emotional dynamics in chat conversations.

\section*{Acknowledgment}
This research has been funded by the Science Committee of the Ministry of Science and Higher Education of the Republic of Kazakhstan (Grant No. AP22786412)









\bibliography{sn-bibliography}

\end{document}